\RequirePackage{fix-cm}
\documentclass[natbib,twocolumn]{svjour3}          
\smartqed  
\usepackage{times}
\usepackage{epsfig}
\usepackage{graphicx}
\usepackage{amsmath}
\usepackage{amssymb}
\usepackage{algorithm2e}
\usepackage{algorithmic}
\usepackage{wrapfig}
\usepackage{subfig}

\usepackage{color}
\definecolor{darkgreen}{rgb}{0,.75,0}

\newcommand{\myparagraph}[1]{\noindent{\emph {#1}}}

\newcommand{\fg}{\textsc{\sffamily fg}\xspace}
\newcommand{\fgdo}{\textsc{\sffamily fg+\-do}\xspace}
\newcommand{\fgso}{\textsc{\sffamily fg+\-so}\xspace}
\newcommand{\fggp}{\textsc{\sffamily fg+\-gp}\xspace}
\newcommand{\fggpdoso}{\textsc{\sffamily fg+\-gp+\-do+\-so}\xspace}

\newcommand{\coarse}{\textsc{\sffamily coarse}\xspace}
\newcommand{\coarsegp}{\textsc{\sffamily coarse+\-gp}\xspace}

\def\eg{\emph{e.g. }} 

\def\ie{\emph{i.e. }}

\def\wrt{w.r.t. }

\journalname{International Journal of Computer Vision}
\begin{document}

\title{Towards Scene Understanding with Detailed 3D Object Representations
}


\author{M. Zeeshan Zia \and
	Michael Stark \and
        Konrad Schindler 
}

\institute{M. Zeeshan Zia \at ETH Zurich and Imperial College London\\
		\email{zeeshan.zia@imperial.ac.uk}	
		\and
           Michael Stark \at
              Max Planck Institute for Informatics, Saarbr\"{u}cken, Germany\\
		\email{stark@mpi-inf.mpg.de}
           \and
	   Konrad Schindler \at
              Swiss Federal Institute of Technology (ETH), Zurich \\
              \email{konrad.schindler@geod.baug.ethz.ch}           
}

\date{Received: date / Accepted: date}

\maketitle

\begin{abstract}
Current approaches to semantic image and scene understanding typically
employ rather simple object representations such as 2D or 3D bounding
boxes. While such coarse models are robust and allow for reliable
object detection, they discard much of the information about objects'
3D shape and pose, and thus do not lend themselves well to
higher-level reasoning.
Here, we propose to base scene understanding on a high-resolution
object representation. An object class -- in our case cars --- is
modeled as a deformable 3D wireframe, which enables fine-grained
modeling at the level of individual vertices and faces.
We augment that model to explicitly include vertex-level occlusion,
and embed all instances in a common coordinate frame, in order to
infer and exploit object-object interactions.
Specifically, from a single view we jointly estimate the shapes and
poses of multiple objects in a common 3D frame. A ground plane in that
frame is estimated by consensus among different objects, which
significantly stabilizes monocular 3D pose estimation.
The fine-grained model, in conjunction with the explicit 3D scene
model, further allows one to infer part-level occlusions between the
modeled objects, as well as occlusions by other, unmodeled scene
elements.
To demonstrate the benefits of such detailed object class models in
the context of scene understanding we systematically evaluate our
approach on the challenging KITTI street scene dataset.
The experiments show that the model's ability to utilize image
evidence at the level of individual parts improves monocular 3D pose
estimation w.r.t.\ both location and (continuous) viewpoint.
\end{abstract}

\section{Introduction}
\label{sec:intro}
\begin{figure}[t!]
\centering
\includegraphics[width=1.0\columnwidth]{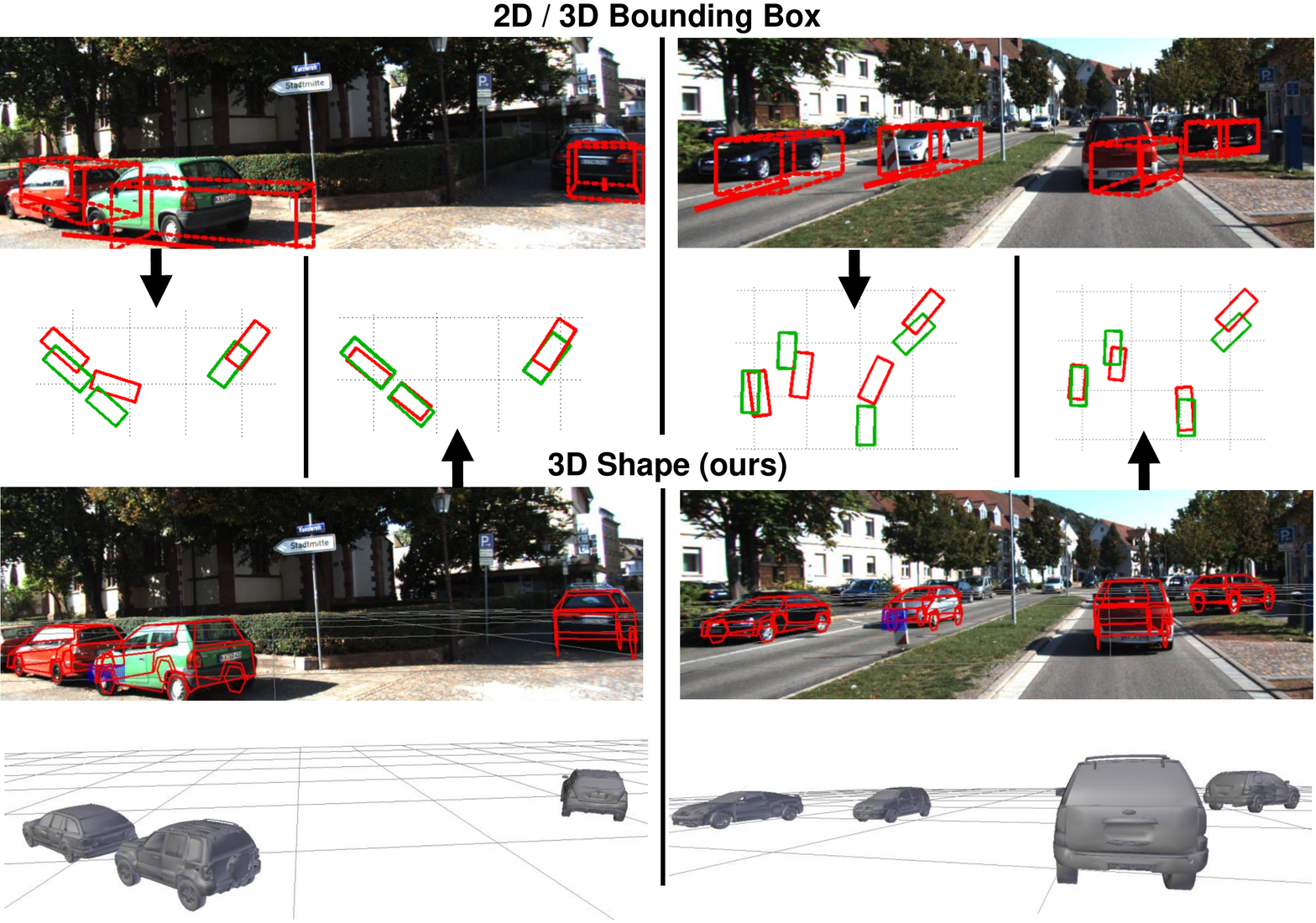}
\caption{{\em Top:} Coarse 3D object bounding boxes
derived from 2D bounding box detections (not shown).  {\em Bottom:}
our fine-grained 3D shape model fits improve 3D localization (see
bird's eye views).}
\label{fig:teaser}
\end{figure}
The last ten years have witnessed great progress in automatic visual
recognition and image understanding, driven by advances in local
appearance descriptors, the adoption of discriminative classifiers, and
more efficient techniques for probabilistic inference. In several
different application domains we now have semantic vision sub-systems
that work on real-world images.
Such powerful tools have sparked a renewed interest in the grand
challenge of visual 3D scene understanding. Meanwhile, individual
object detection performance has reached a plateau after a decade of
steady gains \citep{everingham10ijcv}, further emphasizing the need
for contextual reasoning.

A number of geometrically rather coarse scene-level reasoning systems
have been proposed over the past few years
\citep{hoiem08ijcv,wang10eccv,hedau10eccv,gupta10eccv,silberman12eccv},
which apart from adding more holistic scene understanding also improve
object recognition. The addition of context and the step to reasoning
in 3D (albeit coarsely) makes it possible for different vision
sub-systems to interact and improve each other's estimates, such that
the sum is greater than the parts.

Very recently, researchers have started to go one step further and
increase the level-of-detail of such integrated models, in order to
make better use of the image evidence. Such models learn not only 2D
object appearance but also detailed 3D shape
\citep{xiang12cvpr,hejrati12nips,zia13pami}. The added detail in the representation, 
typically in the form of wireframe meshes learned from 3D CAD models, makes it possible
 to also reason at higher resolution: beyond measuring image evidence at the level of 
individual vertices/parts one can also handle relations between parts, e.g.\ shape 
deformation and part-level occlusion \citep{zia13cvpr}.
Initial results are encouraging. It appears that the more detailed
scene interpretation can be obtained at a minimal penalty in terms of
robustness (detection rate), so that researchers are beginning to
employ richer object models to different scene understanding
tasks \citep{choi13cvpr,delpero13cvpr,zhao13cvpr,xiang133drr,zia14cvpr}.

Here we describe one such novel system for scene understanding based
on monocular images. Our focus lies on exploring the potential of
jointly reasoning about multiple objects in a common 3D frame, and the
benefits of part-level occlusion estimates afforded by the detailed
representation.
We have shown in previous work \citep{zia13pami} how a detailed 3D
object model enables a richer pseudo-3D ($x,y,scale$) interpretation
of simple scenes dominated by a single, unoccluded object---including
fine-grained categorization, model-based segmentation, and monocular
reconstruction of a ground plane.
Here, we lift that system to true 3D, i.e.\ CAD models are scaled to
their true dimensions in world units and placed in a common, metric 3D
coordinate frame. This allows one to reason about geometric
constraints between multiple objects as well as mutual occlusions, at
the level of individual wireframe vertices.

\myparagraph{Contributions. }
We make the following contributions.

\emph{First}, we propose a viewpoint-invariant method for 3D reconstruction (shape and pose estimation)
 of severely occluded objects in single-view images. To obtain a
complete framework for detection and reconstruction, the novel method
is bootstrapped with a variant of the poselets framework
\citep{bourdev09iccv} adapted to the needs of our 3D object model.

\emph{Second}, we reconstruct scenes consisting of multiple such objects, each with their individual shape
 and pose, in a single inference framework, including geometric constraints between them in the form of a 
common ground plane. Notably, reconstructing the fine detail of each object
also improves the 3D pose estimates (location as well as viewpoint)
for entire objects over a 3D bounding box baseline
(Fig.~\ref{fig:teaser}).

\emph{Third}, we leverage the rich detail of the 3D representation
for occlusion reasoning at the individual vertex level, combining
(deterministic) occlusion by other detected objects with a
(probabilistic) generative model of further, unknown occluders. Again,
integrated scene understanding yields improved 3D
localization compared to independently estimating occlusions for
each individual object.

And \emph{fourth}, we present a systematic experimental study on the
challenging KITTI street scene dataset \citep{geiger12cvpr}. While our
fine-grained 3D scene representation can not yet compete with
technically mature 2D bounding box detectors in terms of recall, it
offers superior 3D pose estimation, correctly localizing $>43\%$ of
the detected cars up to 1$\,$m and $>55\%$ up to 1.5$\,$m, even when
they are heavily occluded.

Parts of this work appear in two preliminary conference papers \citep{zia13cvpr,zia14cvpr}.
The present paper describes our approach in more detail, extends the experimental analysis, 
and describes the two contributions (extension of the basic model to occlusions, respectively scene
 constraints) in a unified manner.

The remainder of this paper is structured as follows.
Sec. 2 reviews related work. Sec. 3 introduces our
3D geometric object class model, extended in Sec. 4 to entire scenes. Sec. 5 gives experimental results,
 and Sec. 6 concludes the paper.

\section{Related Work}
\label{sec:related}
\myparagraph{Detailed 3D object representations. }
Since the early days of computer vision research, detailed and complex
models of object geometry were developed to solve object recognition
in general settings, taking into account viewpoint, occlusion,
and intra-class variation.
Notable examples include the works of \citet{kanade80} and \citet{malik87ijcv},
who lift line drawings of 3D objects by classifying the lines and their intersections to 
common occurring configurations; 
and the classic works of \citet{brooks81} and \citet{pentland86}, who represent complex
objects by combinations of atomic shapes, generalized cones and
super-quadrics. Matching CAD-like models to image edges also made it
possible to address partially occluded objects \citep{lowe87ai} and
intra-class variation \citep{sullivan95cbv}.

Unfortunately, such systems could not robustly handle real world
imagery, and largely failed outside controlled lab environments.
In the decade that followed researchers moved to simpler models,
sacrificing geometric fidelity to robustify the matching of the models
to image evidence---eventually reaching a point where the
best-performing image understanding methods were on one hand
bag-of-features models without any geometric layout, and on the other
hand object templates without any flexibility (largely thanks to
advances in local region descriptors and statistical learning).

However, over the past years researchers have gradually started to
re-introduce more and more geometric structure in object class models
and improve their
performance \citep[\eg][]{leibe06ism,felzenszwalb09pami}. At present
we witness a trend to take the idea even further and revive highly
detailed deformable wireframe
models~\citep{zia09icar,li11pami,zia13pami,xiang12cvpr,hejrati12nips}.
In this line of work, object class models are learnt from either 3D
CAD data \citep{zia09icar, zia13pami} or
images \citep{li11pami}. Alternatively, objects are represented as
collections of planar segments \citep[also learnt from CAD
models,][]{xiang12cvpr} and lifted to 3D with non-rigid
structure-from-motion.
In this paper, we will demonstrate that such fine-grained modelling
also better supports scene-level reasoning.

\myparagraph{Occlusion modeling. }
While several authors have investigated the problem of occlusion in recent
years, little work on occlusions exists for detailed part-based 3D
models, notable exceptions being \citep{li11pami,hejrati12nips}.

Most efforts concentrate on 2D bounding box
detectors in the spirit of
HOG \citep{dalal05cvpr}. \citet{fransens06cvpr} model occlusions with
a binary visibility map over a fixed object window and
infer the map with expectation-maximization. In a similar fashion,
sub-blocks that make up the window descriptor are sometimes classified
into occluded and non-occluded
ones \citep{wang09iccv,gao11cvpr,kwak11iccv}. \citet{vedaldi09nips}
use a structured output model to explicitly account for truncation at
image borders and predict a truncation mask at both training and test
time.
If available, motion \citep{enzweiler10cvpr} and/or
depth \citep{meger11bmvc} can serve as additional cues to determine
occlusion, since discontinuities in the depth and motion fields are
more reliable indicators of occlusion boundaries than texture edges.

Even though quite some effort has gone into occlusion invariance for
global object templates, it is not surprising that part-based models
have been found to be better suited for the task. In fact even fixed
windows are typically divided into regular grid cells that one could
regard as ``parts" \citep{wang09iccv,gao11cvpr,kwak11iccv}.
More flexible models include dedicated DPMs for commonly occuring
object-object occlusion cases \citep{tang12bmvc} and variants of the
extended DPM formulation \citep{girshick11nips}, in which an occluder
is inferred from the absence of part evidence.
Another strategy is to learn a very large number of partial
configurations (``poselets") through clustering \citep{bourdev09iccv},
which will naturally also include frequent occlusion patterns.
The most obvious manner to handle occlusion in a proper
part-based model is to explicitly estimate the oclusion states of the
individual parts, either via RANSAC-style sampling to find unoccluded
ones \citep{li11pami}, or via local
mixtures \citep{hejrati12nips}. Here we also store a binary occlusion
flag per part, but explicitly enumerate allowable occlusion patterns
and restrict the inference to that set.

\myparagraph{Qualitative scene representations. }
Beyond detailed geometric models of individual objects, early
computer vision research also attempted to model entire scenes in 3D
with considerable detail. In fact the first PhD thesis in computer vision \citep{roberts65phd}
 modeled scenes comprising of polyhedral objects, considering self-occlusions as well as 
combining multiple simple shapes to obtain complex objects. \citet{koller93ijcv} used simplified 3D
models of multiple vehicles to track them in road scenes,
whereas \citet{haag99ijcv} included scene elements such as trees and
buildings, in the form of polyhedral models, to estimate their shadows
falling on the road, as well as vehicle motion and
illumination. 

Recent work has revisited these ideas at the level of plane- and
box-type models. E.g.,~\citet{wang10eccv} estimate the geometric layout
of walls in an indoor setting, segmenting out the
clutter. Similarly, \citet{hedau10eccv} estimate the layout of a room
and reason about the locations of the bed as a box in the room. For
indoor settings it has even been attempted to recover physical support
relations, based on RGB-D data \citep{silberman12eccv}. 
For fairly generic outdoor scenes, physical support, volumetric
constraints and occlusions have been included, too, still using boxes
as object models \citep{gupta10eccv}. Also for outdoor images, \citet{liu14cvpr}
partition single views into a set of oriented surfaces,
driven by grammar rules for neighboring segments.
It has also been observed that object detections carry information
about 3D surface orientations, such that they can be jointly estimated
even from a single image \citep{hoiem08ijcv}. Moreover, recent
work suggests that object detection can be improved if one includes
the density of common poses between neighboring object
instances \citep{oramas13iccv}.

All the works indicate that even coarse 3D reasoning allows one to
better guess the (pseudo-)3D layout of a scene, while at the same time
improving 2D recognition. Together with the above-mentioned strength
of fine-grained shape models when it comes to occlusion and viewpoint,
this is in our view a compelling reason to add 3D contextual
constraints also to those fine-grained models.

\myparagraph{Quantitative scene representations. }
A different type of methods also includes scene-level reasoning, but
is tailored to specific applications and is more quantitative in
nature. Most works in this direction target autonomous navigation,
hence precise localization of reachable spaces and obstacles is
important. Recent works for the autonomous driving scenario
include: \citep{ess09pami}, in which multi-pedestrian tracking is done
in 3D based on stereo video, and \citep{geiger11nips,wojek13pami},
both aiming for advanced scene understanding including multi-class
object detection, 3D interaction modeling, as well as semantic
labeling of the image content, from monocular input.
Viewpoint estimates from semantic recognition can also be combined
with interest point detection to improve camera pose and scene
geometry even across wide baselines \citep{bao11cvpr}.

For indoor settings, a few recent papers also employ detailed object
representations to support scene understanding \citep{delpero13cvpr},
try to exploit frequently co-occurring object poses \citep{choi13cvpr},
and even supplement geometry and appearance constraints with
affordances to better infer scene layout \citep{zhao13cvpr}.
\section{3D Object Model}
\label{sec:object}
\begin{figure}[t]
\includegraphics[width=1.0\columnwidth]{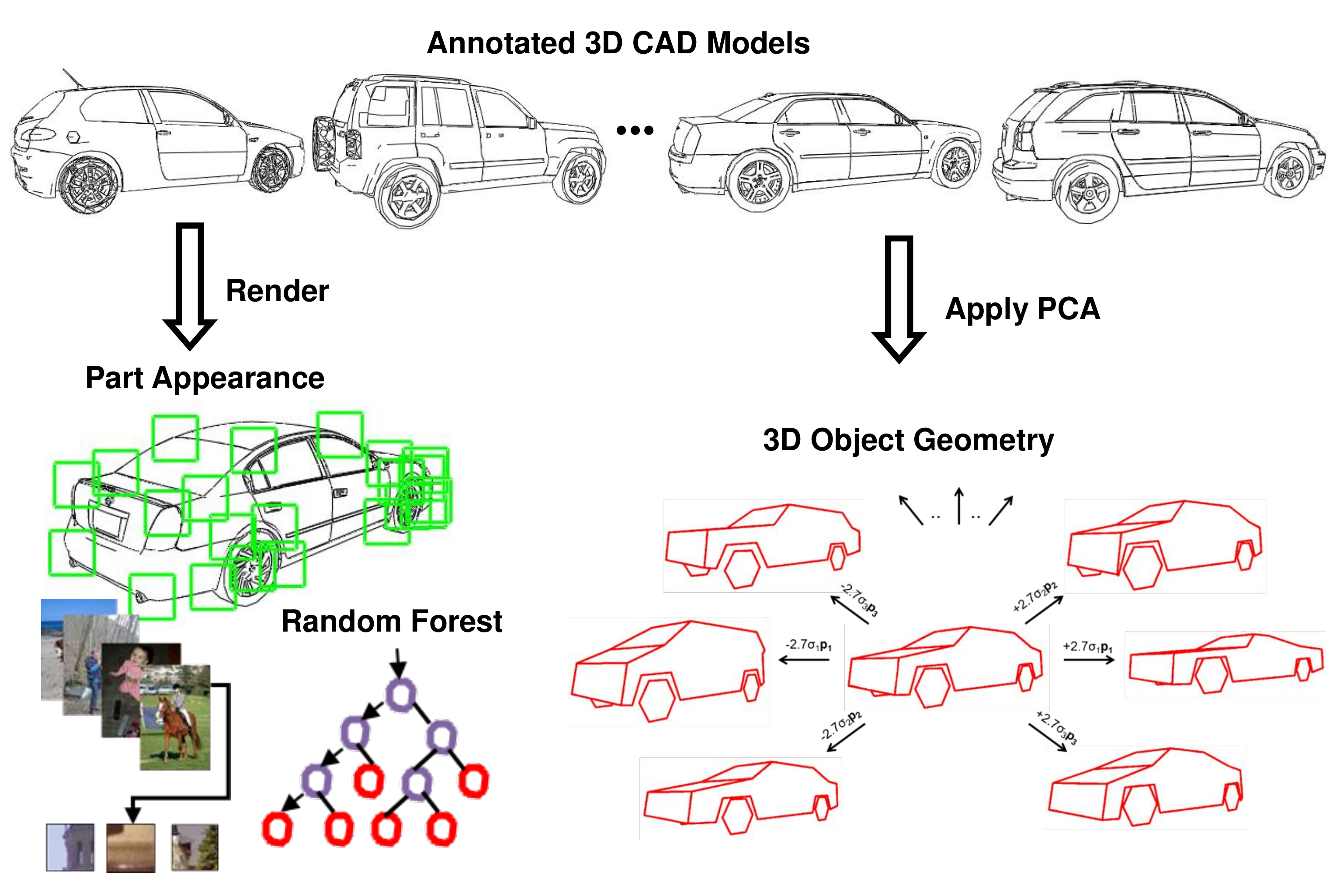}
\caption{3D Object Model.}
\label{fig:ob_illustration}
\end{figure}

We commence by introducing the fine-grained 3D object model that lies
at the core of our approach. Its extension to entire multi-object
scenes will be discussed in Sec.~\ref{sec:scene}. By modeling an
object class at the fine level of detail of individual wireframe
vertices the object model provides the basis for reasoning about
object extent and occlusion relations with high fidelity. To that end,
we lift the pseudo-3D object model that we developed
in~\citet{zia13pami} to metric 3D space, and combine it with the
explicit representation of likely occlusion patterns
from~\citet{zia13cvpr}. Our object representation then comprises a
model of global object geometry (Sec.~\ref{sec:geommodel}), local part
appearance (Sec.~\ref{sec:parts}), and an explicit representation of
occlusion patterns (Sec.~\ref{sec:occluder}). Additionally, the object
representation also includes a grouping of local parts into semi-local
part configurations (Sec.~\ref{sec:partconfigurations}), which will be
used to initialize the model during inference (Sec.~\ref{sec:infer}).
We depict the 3D object representation in Fig.~\ref{fig:ob_illustration}.

\subsection{Global Object Geometry}
\label{sec:geommodel}

We represent an object class as a deformable 3D wireframe, as in
the classical ``active shape model"
formulation~\citep{cootes95cviu}. The vertices of the wireframe are
defined manually, and wireframe exemplars are collected by annotating
a set of 3D CAD models (i.e., selecting corresponding vertices from
their triangle meshes). Principal Component Analysis (PCA) is applied
to obtain the mean configuration of vertices in 3D as well as the
principal modes of their relative displacement.  The final geometric
object model then consists of the mean wireframe $\boldsymbol\mu$ plus
the $m$ principal component directions $\mathbf{p}_j$ and
corresponding standard deviations $\sigma_j$, where $1 \leq j \leq m$.
Any 3D wireframe $\mathbf{X}$ can thus be represented, up to some
residual $\boldsymbol\epsilon$, as a linear combination of $r$
principal components with geometry parameters $\mathbf{s}$, where
$s_k$ is the weight of the $k^{th}$ principal component:
\begin{equation}
\mathbf{X}(\mathbf{s})
= \boldsymbol\mu + \sum_{k=1}^r s_k \sigma_k \mathbf{p}_k + \boldsymbol\epsilon
\label{eqn:pcamodel}
\end{equation}
Unlike the earlier~\cite{zia13pami}, the 3D CAD models are scaled
according to their real world metric dimensions. %
\footnote{While in the earlier work they were scaled to the same size, so as to keep
 the deformations from the mean shape small.}%
The resulting metric PCA model hence encodes physically meaningful
scale information in world units, that allow one to assign absolute 3D
positions to object hypotheses (given known camera intrinsics).
%

\subsection{Local Part Appearance}
\label{sec:parts}

We establish the connection between the 3D geometric object model
(Sec.~\ref{sec:geommodel}) and an image by means of a set of
\emph{parts}, one for each wireframe vertex. For each part, a multi-view
appearance model is learned, by generating from training patches with
non-photorealistic rendering of 3D CAD models from a large number of
different viewpoints~\citep{stark10bmvc}, and training a
sliding-window detector on these patches.

Specifically, we encode patches around the projected locations of the
annotated parts ($\approx\,$10\% in size of the full object width) as
dense shape context features~\citep{belongie00nips}. We learn a
multi-class Random Forest classifier where each class represents the
multi-view appearance of a particular part. We also dedicate a class
trained on background patches, combining random real image patches
with rendered non-part patches to avoid classifier bias. Using
synthetic renderings for training allows us to densely sample the
relevant portion of the viewing sphere with minimal annotation effort
(one time labeling of part locations on 3D CAD models, i.e.\ no added
effort in creating the shape model).

\subsection{Explicit Occluder Representation}
\label{sec:occluder}

The 3D wireframe model allows one to represent partial occlusion at
the level of individual parts: each part has an associated binary
variable that stores whether the part is visible or occluded. Note
that, in theory, this results in a exponential number of possible
combinations of occluded and unoccluded parts, hindering efficient
inference over occlusion states. We therefore take advantage of the
fact that partial occlusion is not entirely random, but tends to
follow re-occurring patterns that render certain joint occlusion
states of multiple parts more likely than others~\citep{pepik13cvpr}:
the joint occlusion state depends on the shape of the occluding
physical object(s).

\begin{figure}[t]
\centering
\includegraphics{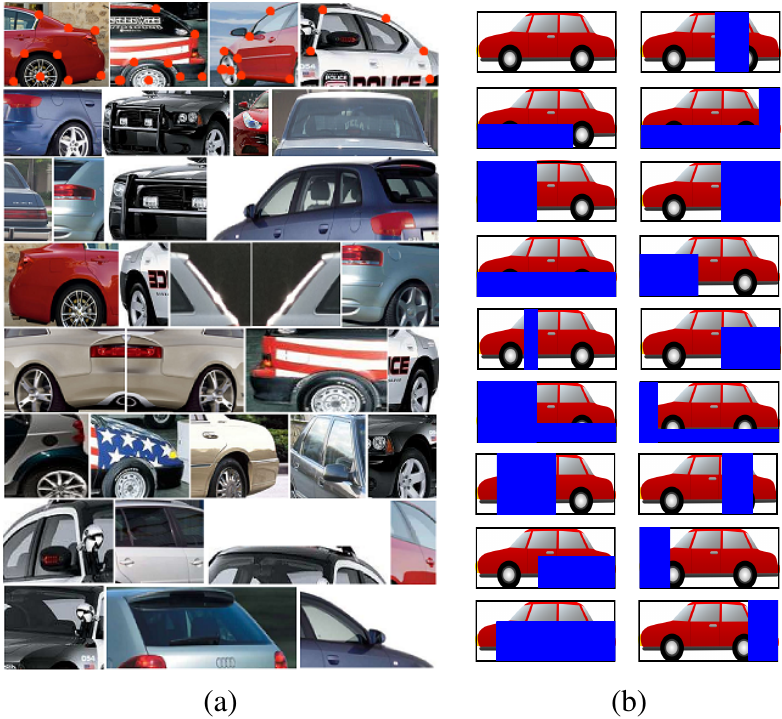}
\caption{(a) Individual training examples for a few part $configurations$ (top row shows labeled part locations), (b) example occlusion masks.}
\label{fig:occmasks_poselet}
\end{figure}

Here we approximate the shapes of (hypothetical) occluders as a finite
set of occlusion masks, following~\citep{kwak11iccv,zia13cvpr}. This
set of masks constitutes a (hard) non-parameteric prior over possible
occlusion patterns. The set is denoted by $\{a_i\}$, and for
convenience we denote the empty mask which leaves the object fully
visible by $a_0$.
We sample the set of occlusion masks regularly from
a generative model, by sliding multiple boxes across the mask in small
spatial increments (the parameters of those boxes are determined
empirically). Figure~\ref{fig:occmasks_poselet}(b) shows a few out of
the total 288 masks in our set, with the blue region representing the
occluded portion of the object (car). The collection is able to
capture different modes of occlusion, for example truncation by the
image border (Fig.~\ref{fig:full_vs_coarse}(d), first row), occlusion
in the middle by a post or tree (Fig.~\ref{fig:full_vs_coarse}(d), 2nd
row), or occlusion of only the lower parts from one side
(Fig.~\ref{fig:full_vs_coarse}(d), third row).

Note that the occlusion mask representation is independent of the
cause of occlusion, and allows to uniformly treat occlusions that
arise from {\em (i)} self occlusion (a part is occluded by a wireframe
face of the same object), {\em (ii)} occlusion by another object that
is part of the same scene hypothesis (a part is occluded by a
wireframe face of another object), {\em (iii)} occlusion by an unknown
source (a part is occluded by an object that is not part of the same
scene hypothesis, or image evidence is missing).
%

\subsection{Semi-Local Part Configurations}
\label{sec:partconfigurations}

In the context of people detection and pose estimation, it has been
realized that individual body parts are hard to accurately localize,
because they are small and often not discriminative enough in
isolation~\citep{bourdev09iccv}. Instead, it has proved beneficial to
train detectors that span multiple parts appearing in certain poses
(termed ``poselets''), seen from a certain viewpoint, and selecting
the ones that exhibit high discriminative power against background on
a validation set (alternately, the scheme of \citet{maji09cvpr} could also be used).
In line with these findings, we introduce the notion of part {\em
configurations}, i.e.\ semi-local arrangements of a number of parts, 
seen from a specific viewpoint, that are adjacent (in terms of
wireframe topology). Some examples are depicted in
Fig.~\ref{fig:occmasks_poselet}(a)). These configurations provide more
reliable evidence for each of the constituent parts than individual
detectors. We use detectors for different configurations to find
primising 2D bounding boxes and viewpoint estimates, as
initializations for fitting the fine-grained 3D object models.

Specifically, we list all the possible configurations of 3-4 adjacent visible 
parts that are not smaller than $\approx 20\%$ of the full object (for the eight coarse viewpoints).
 Some configurations cover the full
car, whereas others only span a part of it (down to $\approx 20\%$ of
the full object). However we 
found the detection performance to be rather consistent even if other heuristics were used for part
configuration generation.  We then train a bank of single component
DPM detectors, one for each configuration, in order to ensure high
recall and a large number of object hypotheses to choose from. At test
time, activations of these detectors are merged together through
agglomerative clustering to form full object hypothesis, in the spirit
of the poselet framework~\citep{bourdev09iccv}.
For training, we utilize a set of images labeled at the level of
individual parts, and with viewpoint labels from a small discrete set
(in our experiments $8$ equally spaced viewpoints).  All the objects
in these images are fully visible. Thus, we can store the relative
scale and bounding box center offsets, \wrt the full object bounding
box, for the part-configuration examples. When detecting potentially
occluded objects in a test image, the activations of all configuration
detectors predict a full object bounding box and a (discrete) pose.

Next we recursively merge nearby (in $x,y,scale$) activations that
have the same viewpoint. Merging is accomplished by averaging the
predicted full object bounding box corners, and assigning it the
highest of the detection scores. After this agglomerative clustering
has terminated all clusters above a fixed detection score are picked
as legitimate objects. Thus we obtain full object bounding box
predictions (even for partially visible objects), along with an
approximate viewpoint.
\section{3D Scene Model}
\label{sec:scene}
We proceed by extending the single object model of
Sec.~\ref{sec:object} to entire scenes, where we can jointly reason
about multiple objects and their geometric relations, placing them on
a common ground plane and taking into account mutual occlusions. As we
will show in the experiments (Sec.~\ref{sec:experiments}), this joint
modeling can lead to significant improvements in terms of 3D object
localization and pose estimation compared to separately modeling
individual objects. It is enabled by a joint scene hypothesis space
(Sec.~\ref{sec:scenehypo}), governed by a probabilistic formulation
that scores hypotheses according to their likelihood
(Sec.~\ref{sec:probform}), and an efficient approximate inference
procedure for finding plausible scenes (Sec.~\ref{sec:infer}).
The scene model is schematically depicted in Fig.~\ref{fig:illustration}.

\subsection{Hypothesis Space}
\label{sec:scenehypo}
\begin{figure}[t]
\includegraphics[width=1.0\columnwidth]{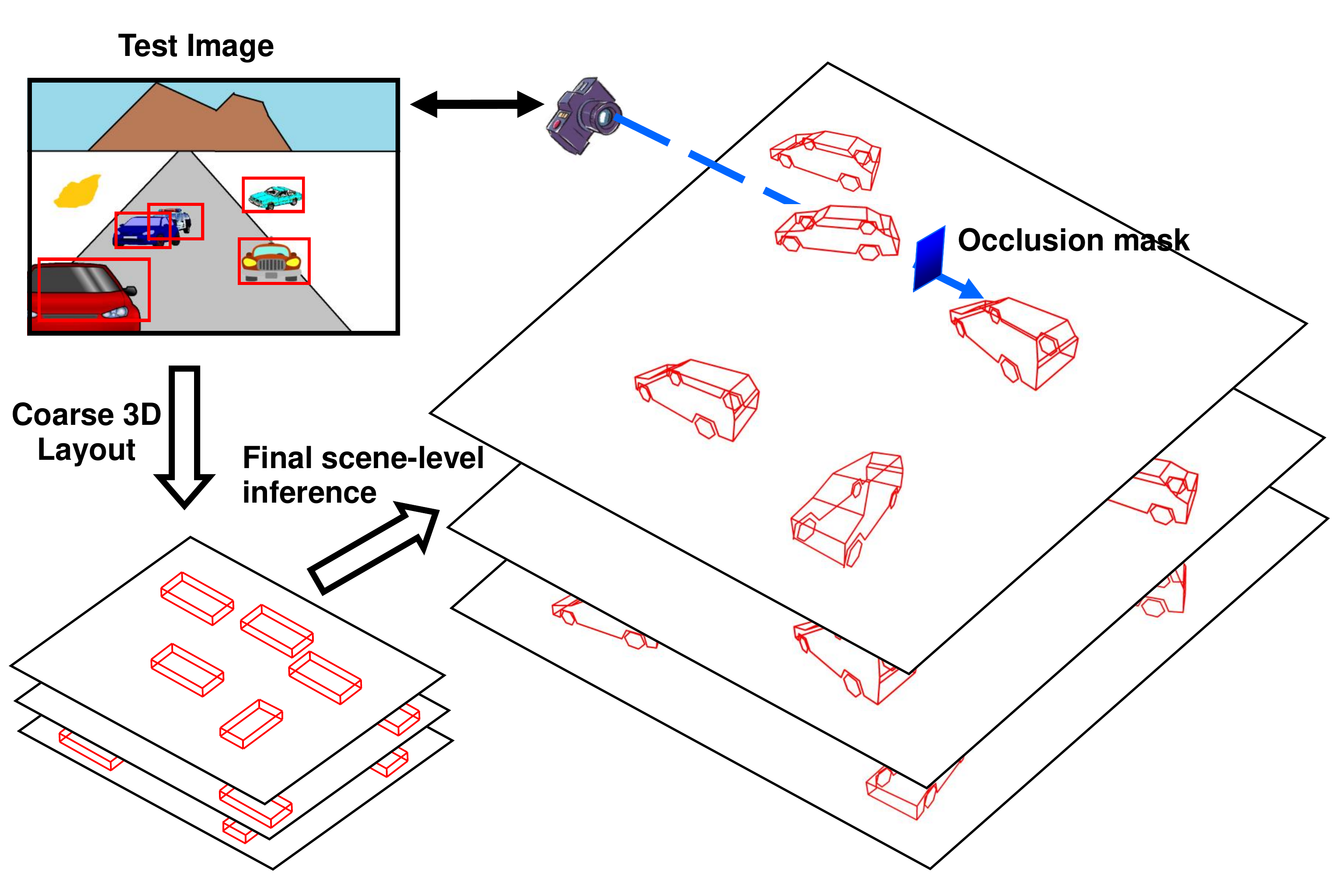}
\caption{3D Scene Model.}
\label{fig:illustration}
\end{figure}
Our 3D scene model comprises a common ground plane and a set of 3D
deformable wireframes with corresponding occlusion masks
(Sec.~\ref{sec:object}). Note that this hypothesis space is more
expressive than the 2.5 D representations used by previous work
\citep{ess09pami,meger11bmvc,wojek13pami}, as it allows reasoning
about locations, shapes, and interactions of objects, at the level of
individual 3D wireframe vertices and faces.

\myparagraph{Common ground plane.}
In the full system, we constrain all the object instances to lie on a
common ground plane, as often done for street scenes. This assumption
usually holds and drastically reduces the search space for possible
object locations ($2$ degrees of freedom for translation and $1$ for
rotation, instead of $3+3$). Moreover, the consensus for a common
ground plane stabilizes 3D object localization.
%
%
We parametrize the ground plane with the pitch and roll angles
relative to the camera frame, $\boldsymbol{\theta}_{gp} =
(\theta_{pitch}, \theta_{roll}$). The height $q_y$ of the camera above
ground is assumed known and fixed.

\myparagraph{Object instances.}
Each object in the scene is an instance of the 3D wireframe model
described in Sec.~\ref{sec:geommodel}. An individual instance
$\mathbf{h^\beta} = (\mathbf{q}, \mathbf{s}, a)$ comprises 2D
translation and azimuth $\mathbf{q}=(q_x,q_z,q_{az})$ relative to the
ground plane, shape parameters $\mathbf{s}$, and an occlusion mask
$a$.
%

\myparagraph{Explicit occlusion model.}
As detailed in Sec.~\ref{sec:occluder}, we represent occlusions on
an object instance by selecting an occluder mask out of a pre-defined
set $\{a_i\}$, which in turn determines the binary occlusion state of
all parts. That is, the occlusion state of part $j$ is given by an
indicator function
$o_j( \boldsymbol{\theta}_{gp},q_{az},\mathbf{s}, a)$, with
$\boldsymbol{\theta}_{gp}$ the ground plane parameters, $q_{az}$ the
object azimuth, $\mathbf{s}$ the object shape, and $a$ the occlusion
mask. 
Since all object hypotheses reside in the same 3D coordinate system,
mutual occlusions can be derived deterministically from their depth
ordering (Fig.~\ref{fig:illustration}): we cast rays from the camera
center to each wireframe vertex of all other objects, and record
intersections with faces of any other object as an appropriate
occlusion mask. Accordingly, we write
$\Gamma\big(\{\mathbf{h^{1}},\mathbf{h^{2}},\ldots,\mathbf{h^{n}} \}\backslash\mathbf{h^\beta},\mathbf{h^\beta},\boldsymbol{\theta}_{gp}\big)$,
\ie\ the operator $\Gamma$ returns the index of the occlusion mask for
$\mathbf{h^\beta}$ as a function of the other objects in a given scene
estimate.

\subsection{Probabilistic Formulation}
\label{sec:probform}
All evidence in our model comes from object part detection, and the
prior for allowable occlusions is given by per-object occlusion masks and relative object positions
(Sec.~\ref{sec:scenehypo}).


\myparagraph{Object likelihood.}
The likelihood of an object being present at a particular location in
the scene is measured by responses of a bank of
(viewpoint-independent) sliding-window part detectors
(Sec.~\ref{sec:parts}), evaluated at projected image coordinates of
the corresponding 3D wireframe vertices.%
\footnote{In practice this amounts to a look-up in the precomputed response maps.} %
The likelihood
$\mathcal{L}(\mathbf{h^\beta}, \boldsymbol{\theta}_{gp})$ for an
object $\mathbf{h^\beta}$ standing on the ground plane
$\boldsymbol{\theta}_{gp}$ is the sum over the responses of all
visible parts, with a constant likelihood for occluded parts ($m$ is the 
total number of parts, $a_0$ is the 'full visibility' occluder mask):
\begin{equation}
\!\mathcal{L}(\mathbf{h^\beta}, \boldsymbol{\theta}_{gp})\!=\!\max_{\boldsymbol{\varsigma}} \Bigg[\!\frac{\sum_{j=1}^{m}\!\big( \mathcal{L}_v+\mathcal{L}_o \big)}{\sum_{j=1}^{m}o_j(\boldsymbol{\theta}_{gp},q_{az},\mathbf{s}, a_0)}\Bigg].
\end{equation}
The denominator normalizes for the varying number of self-occluded
parts at different viewpoints. $\mathcal{L}_v$ is the evidence (pseudo
log-likelihood) $S_j(\boldsymbol{\varsigma},\mathbf{x}_j)$ for part
$j$ if it is visible, found by looking up the detection score at image
location $\mathbf{x}_j$ and scale $\boldsymbol{\varsigma}$, normalized
with the background score $S_b ( \boldsymbol{\varsigma},\mathbf{x}_j)$
as in~\citep{villamizar11bmvc}.
$\mathcal{L}_o$ assigns a fixed likelihood $c$, estimated by cross-validation on a held-out dataset:
\begin{align}
&\mathcal{L}_v = o_j(\boldsymbol{\theta}_{gp},q_{az},\mathbf{s}, a) \log\frac{S_j(\boldsymbol{\varsigma},\mathbf{x}_j)} {S_b(\boldsymbol{\varsigma},\mathbf{x}_j)}\;,
\\
&
\mathcal{L}_{o} =
\big(o_j(\boldsymbol{\theta}_{gp},q_{az},\mathbf{s}, a_0)
- o_j(\boldsymbol{\theta}_{gp},q_{az},\mathbf{s}, a)\big) c\;.
\label{eqn:occluterm}
\end{align}

\myparagraph{Scene-level likelihood.}
To score an entire scene we combine object hypotheses and ground plane
into a scene hypothesis $\mathbf{\psi} =
\{q_y,{\boldsymbol{\theta}_{gp}},\mathbf{h^1},...,\mathbf{h^n}\}$.
The likelihood of a complete scene is then the sum over all object
likelihoods, such that the objective for scene interpretation becomes:
\begin{align}
\label{eqn:objective}
&
\!\mathbf{\hat{\mathbf{\psi}}} = {\arg \max}_{\psi}
\Bigg[\sum_{\beta=1}^{n}
\mathcal{L}(\mathbf{h^\beta}, \boldsymbol{\theta}_{gp})
\Bigg] \;.
\end{align}
Note, the domain
$Dom\big(\mathcal{L}\big(\mathbf{h^\beta},\boldsymbol{\theta}_{gp})\big)$
must be limited such that the occluder mask $a^\beta$ of an object
hypothesis $\mathbf{h^\beta}$ is dependent on relative poses of
all the objects
in the scene: an object hypothesis $\mathbf{h^\beta}$ can only be
assigned occlusion masks $a_i$ which respect object-object
occlusions---\ie at least all the vertices covered by
$\Gamma\big(\{\mathbf{h^{1}},\mathbf{h^{2}},\ldots,\mathbf{h^{n}} \}\backslash\mathbf{h^\beta},\mathbf{h^\beta},\boldsymbol{\theta}_{gp})\big)$
must be covered, even if a different mask would give a higher objective value.
Also note that the ground plane in our current implementation is a hard
constraint---objects off the ground are impossible in our
parameterization (except for experiments in which we ``turn off" the
ground plane for comparison).

\subsection{Inference}
\label{sec:infer}
The objective function in Eqn.~\ref{eqn:objective} is
high-dimensional, highly non-convex, and not smooth (due to the binary
occlusion states). Note that deterministic occlusion reasoning
potentially introduces dependencies between all pairs of objects, and
the common ground plane effectively ties all other variables to the
ground plane parameters $\boldsymbol{\theta}_{gp}$. In order to still
do approximate inference and reach strong local maxima of the
likelihood function, we have designed an inference scheme that
proceeds in stages, lifting an initial 2D guess
(\emph{Initialization}) about object locations to a coarse 3D model
(\emph{Coarse 3D Geometry}), and refining that coarse model into a
final collection of consistent 3D shapes (\emph{Final scene-level
inference, Occlusion Reasoning}).

\myparagraph{Initialization.}
We initialize the inference from coarse 2D bounding box pre-detections
and corresponding discrete viewpoint estimates
(Sec.~\ref{sec:partconfigurations}), keeping all pre-detections above
a confidence threshold. Note that this implicitly determines the
maximum number of objects that will be considered in the scene
hypothesis under consideration.

\myparagraph{Coarse 3D geometry.}
Since we reason in a fixed, camera-centered 3D coordinate frame, the
initial detections can be directly lifted to 3D space, by casting rays
through 2D bounding box centers and instantiating objects on these
rays, such that their reprojections are consistent with the 2D boxes
and discrete viewpoint estimates, and reside on a common ground plane.
In order to avoid discretization artifacts, we then refine the lifted
object boxes by imputing the mean object shape and performing a grid
search over ground plane parameters and object translation and
rotation (azimuth).
In this step, rather than commiting to a single scene-level hypothesis,
we retain many candidate hypotheses ({\em scene particles}) that are
consistent with the 2D bounding boxes and viewpoints of the
pre-detections within some tolerance.

\myparagraph{Occlusion reasoning.}
We combine two different methods to select an appropriate occlusion
mask for a given object, {\em (i)} deterministic occlusion reasoning,
and {\em (ii)} occlusion reasoning based on (the absence of) part
evidence.

{\em (i)} Since by construction we recover the 3D locations and shapes
of multiple objects in a common frame, we can calculate whether a
certain object instance is occluded by any other modeled object
instance in our scene. This is calculated efficiently by casting rays
to all (not self-occluded) vertices of the object instance, and
checking if a ray intersects any other object in its path before
reaching the vertex. This deterministically tells us which parts of
the object instance are occluded by another modeled object in the
scene, allowing us to choose an occluder mask that best represents the
occlusion (overlaps the occluded parts). To select the best mask we
search through the entire set of occluders to maximize the number of
parts with the correct occlusion label, with greater weight on the
occluded parts (in the experiments, twice as much as for visible parts).

{\em (ii)} For parts not under deterministic occlusion, we look for
missing image evidence (low part detection scores for multiple
adjacent parts), guided by the set of occluder masks. Specifically,
for a particular wireframe hypothesis, we search through the set of
occluder masks to maximize the summed part detection scores (obtained
from the Random Forest classifier, Sec.~\ref{sec:parts}), replacing
the scores for parts behind the occluder by a constant (low) score
$c$. Especially in this step, leveraging local context in the form of
occlusion masks stabilizes individual part-level occlusion estimates,
which by themselves are rather unreliable because of the noisy
evidence.

\myparagraph{Final scene-level inference.}
Finally, we search a good local optimum of the scene objective
function (Eqn.~\ref{eqn:objective}) using an iterative stochastic
optimization scheme shown in Algorithm~\ref{algo:inference}.
Each particle is iteratively refined in two steps: first, the shape
and viewpoint parameters of all objects are updated. Then, object
occlusions are recomputed and occlusions by unmodeled objects are
updated, by exhaustive search over the set of possible masks.

The update of the continuous shape and viewpoint follows the
smoothing-based optimization of~\citet{leordeanu08cvpr}. In a
nutshell, new values for the shape and viewpoint parameters are found
by testing many random perturbations around the current values.
The trick is that the random perturbations follow a normal
distribution that is adapted in a data-driven fashion: in regions
where the objective function is unspecific and wiggly the variance is
increased to suppress weak local minima; near distinct peaks the
variance is reduced to home in on the nearby stronger optimum. For
details we refer to the original publication.

For each scene particle the two update steps -- shape and viewpoint
sampling for all cars with fixed occlusion masks, and exhaustive
occlusion update for fixed shapes and viewpoints -- are iterated, and
the particle with the highest objective value $\psi$ forms our MAP
estimate.
As the space of ground planes is already well-covered by the set of
multiple scene particles (in our experiments 250), we keep the ground
plane parameters of each particle constant. This stabilizes the
optimization.
Moreover, we limit ourselves to a fixed number of objects from the
pre-detection stage. The scheme could be extended to allow adding and
deleting object hypotheses, by normalizing the scene-level likelihood
with the number of object instances under consideration.

\begin{algorithm}[t]
\begin{footnotesize}
\SetInd{0.5em}{0.5em}
%
\textbf{Given:}
Scene particle $\psi^{'}$: initial objects
$\mathbf{h}^\beta=(\mathbf{q}^\beta,\mathbf{s}^\beta,a^\beta)$,\\
$\beta=1\ldots n$; fixed $\boldsymbol{\theta}_{gp}$; $a^\beta = a_0$
(all objects fully visible)\\
\For{fixed number of iterations}
{
\textbf{1. }\For{$\beta=1 \ldots n$}
{
{\bf draw samples}
$\{\mathbf{q}^\beta_j,\mathbf{s}^\beta_j\}^{j=1..m}$ from a
Gaussian\\
$\quad\mathcal{N}(\mathbf{q}^\beta,\mathbf{s}^\beta;\Sigma^\beta)$ centered at current values;\\
{\bf update} $\mathbf{h}^\beta=\text{argmax}_j\;\mathcal{L}\big(\mathbf{h}^\beta(\mathbf{q}^\beta_j,\mathbf{s}^\beta_j,a^\beta),\boldsymbol{\theta}_{gp}\big)$\\
}
\textbf{2. }\For{$\beta=1 \ldots n$}
{
{\bf update} occlusion mask (exhaustive search)
$\quad a^\beta=\text{argmax}_{j}\;\; \mathcal{L}\big(\mathbf{h}^\beta(\mathbf{q}^\beta,\mathbf{s}^\beta,a_j),\boldsymbol{\theta}_{gp}\big)$\\
}
%
\textbf{3. Recompute sampling variance} $\Sigma^\beta$ of Gaussians~\citep{leordeanu08cvpr}\\
}
\end{footnotesize}
\caption{Inference run for each scene particle.}
\label{algo:inference}
\end{algorithm}
\section{Experiments}
\label{sec:experiments}
In this section, we extensively analyze the performance of our
fine-grained 3D scene model, focusing on its ability to derive 3D
estimates from a single input image (with known camera intrinsics). To
that end, we evaluate object localization in 3D metric space
(Sec.~\ref{sec:3d_localization}) as well as 3D pose estimation
(Sec.~\ref{sec:3d_viewpoint}) on the challenging KITTI
dataset~\citep{geiger12cvpr} of street scenes. In addition, we analyze
the performance of our model w.r.t. part-level occlusion prediction
and part localization in the 2D image plane
(Sec.~\ref{sec:evaluate2d}).
In all experiments, we compare the performance of our full model with
stripped-down variants as well as appropriate baselines, to highlight
the contributions of different system components to overall
performance.

\subsection{Dataset}
\label{sec:dataset}
In order to evaluate our approach for 3D layout estimation from a
single view, we require a dataset with 3D annotations. We thus turn to
the KITTI \emph{3D object detection and orientation
estimation} benchmark dataset~\citep{geiger12cvpr} as a testbed for our
approach, since it provides challenging images of realistic street
scenes with varying levels of occlusion and clutter, but nevertheless
controlled enough conditions for thorough evaluations. It consists of
around $7,500$ training and $7,500$ test images of street scenes
captured from a moving vehicle and comes with labeled 2D and 3D object
bounding boxes and viewpoints (generated with the help of a laser
scanner).

\myparagraph{Test set.} Since annotations are only made publicly
available on the training set of KITTI, we utilize a portion of this
training set for our evaluation. We choose only images with multiple
cars that are large enough to identify parts, and manually annotate
all cars in this subset with 2D part locations and part-level
occlusion labels. Specifically, we pick every 5th image from the
training set with at least two cars with height greater than $75$
pixels. This gives us $260$ test images with $982$ cars in total, of
which $672$ are partially occluded, and $476$ are severely
occluded. Our selection shall ensure that while being biased towards
more complex scenes, we still sample a representative portion of the
dataset.

\myparagraph{Training set.} We use two different kinds of data for
training our model, {\em (i)} synthetic data in the form of rendered
CAD models, and {\em (ii)} real-world training data.
{\em (i)} We utilize $38$ commercially available 3D CAD models of cars
for learning the object wireframe model as well as for learning
viewpoint-invariant part appearances, \citep[c.f.][]{zia13pami}.
Specifically, we render the 3D CAD models from $72$ different azimuth
angles ($5^\circ$ steps) and 
$2$ elevation angles ($7.5^\circ$ and $15^\circ$ above the ground),
densely covering the relevant part of the viewing sphere, using the
non-photorealistic style of~\citet{stark10bmvc}. Rendered part patches
serve as positive part examples, randomly sampled image patches as
well as non-part samples from the renderings serve as negative
background examples to train the multi-class Random Forest
classifier. The classifier distinguishes $37$ classes ($36$ parts and
$1$ background class), using $30$ trees with a maximum depth of
$13$. The total number of training patches is $162,000$, split into
$92,000$ part and $70,000$ background patches.
{\em (ii)} We train $118$ part configuration detectors (single
component DPMs) labeled with discrete viewpoint, 2D part locations and
part-level occlusion labels on a set of $1,000$ car images downloaded
from the internet and $150$ images from the KITTI dataset (none of
which are part of the test set).
In order to model the occlusions, we semi-automatically
define a set of $288$ occluder masks, the same as
in~\citet{zia13cvpr}.

\subsection{Object Pre-Detection}
\label{sec:predet}
As a sanity check, we first verify that our 2D pre-detection
(Sec.~\ref{sec:partconfigurations}) matches
the state-of-the-art. To that end we evaluate a standard 2D bounding
box detection task according to the PASCAL VOC
criterion ($>50\%$ intersection-over-union
between predicted and ground truth bounding boxes).
As normally done we restrict the evaluation to objects of a certain
minimum size and visibility. Specifically, we only consider cars
$>50$ pixels in height which are at least $20\%$ visible. The
minimum size is slightly stricter than the $40$ pixels
that~\citet{geiger12cvpr} use for the dataset (since we need to ensure
enough support for the part detectors), whereas the occlusion
threshold is much more lenient than their $80\%$ (since we are
specifically interested in occluded objects).

\begin{figure}[thb]
\centering
\includegraphics[width=0.6\columnwidth]{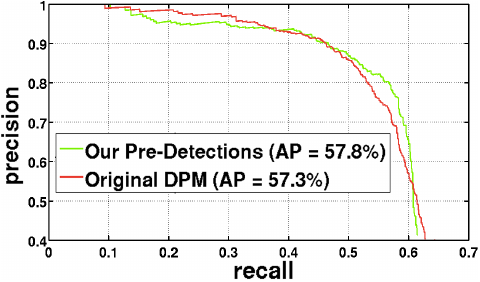}
\caption{Object pre-detection performance.}
\label{fig:predetection}
\end{figure}

\myparagraph{Results.}
We compare our bank of single component DPM detectors to the original
deformable part model~\citep{felzenszwalb09pami}, both trained on the
same training set (Sec.~\ref{sec:dataset}).  Precision-recall curves
are shown in Fig.~\ref{fig:predetection}. We observe that our detector
bank (green curve, $57.8\%$ AP) in fact performs slightly better than
the original DPM (red curve, $57.3\%$ AP). In addition, it delivers
coarse viewpoint estimates and rough part locations that we can
leverage for initializing our scene-level inference
(Sec.~\ref{sec:infer}). The pre-detection takes about 2
minutes per test image on a single core (evaluation of 118 single
component DPMs and clustering of their votes).

\subsection{Model Variants and Baselines}
\label{sec:variants}
We compare the performance of our full system with a number of stripped
down variants in order to quantify the benefit that we get from each
individual component. We consider the following variants:

\emph{(i)}~\fg: the basic version of our fine-grained 3D object
model, without ground plane, searched occluder or deterministic
occlusion reasoning; this amounts to independent modeling of the
objects in a common, metric 3D scene coordinate system.
\emph{(ii)}~\fgso: same as (i) but with searched occluder
to represent occlusions caused by unmodeled scene elements.
\emph{(iii)}~\fgdo: same as (i) but with deterministic occlusion 
reasoning between multiple objects.
\emph{(iv)}~\fggp: same as (i), but with common ground plane.
\emph{(v)}~\fggpdoso: same as (i), but with all three components,
common ground plane, searched occluder, and deterministic occlusion
turned on.
\emph{(vi)} the earlier pseudo-3D shape model~\citep{zia13cvpr}, with
probabilistic occlusion reasoning; this uses essentially the same
object model as (ii), but learns it from examples scaled to
the \emph{same} size rather than the \emph{true} size, and fits the model in
2D $(x,y,scale)$-space rather explicitly recovering a 3D scene
interpretation.

We also compare our representation to two different
baselines, \emph{(vii)}~\coarse: a scene model consisting of 3D
bounding boxes rather than detailed cars, corresponding to the coarse
3D geometry stage of our pipeline (Sec.~\ref{sec:infer});
and \emph{(viii)}~\coarsegp: like \emph{(vii)} but with a common ground
plane for the bounding boxes. Specifically, during the coarse grid search we choose 
the 3D bounding box hypothesis whose 2D projection is closest to the corresponding 
pre-detection 2D bounding box.

\subsection{3D Evaluation}
\label{sec:evaluate3d}
%
\begin{table}[t]
\begin{center}
\footnotesize
\renewcommand{\tabcolsep}{1.5pt}
\begin{tabular}{|l|c|c|c|c|c|c|c|c|c|}
\hline
{} & \multicolumn{2}{|c|}{full dataset} & \multicolumn{2}{|c|}{occ \textgreater 0 parts} & \multicolumn{2}{|c|}{occ \textgreater 3 parts}\\
\centering {} & {\textless 1m} & {\textless 1.5m} & {\textless 1m} & {\textless 1.5m} & {\textless 1m} & {\textless 1.5m} \\ 
\hline
\centering {{\em Fig.~\ref{fig:plot_3dloc1} plot}} & {\em (a)} & {\em (b)} &  & & {\em (c)} & {\em (d)} \\
\hline
\hline
{\emph{(i)} \fg} & 23\% & 35\% & 22\% & 31\% & 23\% & 32\%\\ 
{\emph{(ii)} \fgso} & 26\% & 37\% & 23\% & 33\% & 27\% & 36\%\\ 
{\emph{(iii)} \fgdo} & 25\% & 37\% & 26\% & 35\% & 27\% & 38\%\\ 
{\emph{(iv)} \fggp} & 40\% & 53\% & 40\% & 52\% & 38\% & 49\%\\ 
{\emph{(v)} \fggpdoso} & {\bf 44\%} & {\bf 56\%} & {\bf 44\%} & {\bf 55\%} & {\bf 43\%} & {\bf 60\%}\\ 
\hline
{\emph{(vi)} \cite{zia13cvpr}} &  --- & --- & --- &  --- & --- & ---\\
\hline
\hline
{\emph{(vii)} \coarse} & 21\% & 37\% & 21\% & 40\% & 20\% & 42\%\\
{\emph{(viii)} \coarsegp} & 35\% & 54\% & 28\% & 48\% & 27\% & 47\%\\
\hline
\end{tabular}
\end{center}
\caption{3D localization accuracy: percentage of cars correctly
  localized within 1 and 1.5 meters of ground truth.}
\label{tab:results_3d}
\end{table}
%
\begin{figure}[t]
\footnotesize
\begin{center}
\renewcommand{\tabcolsep}{1pt}
\includegraphics{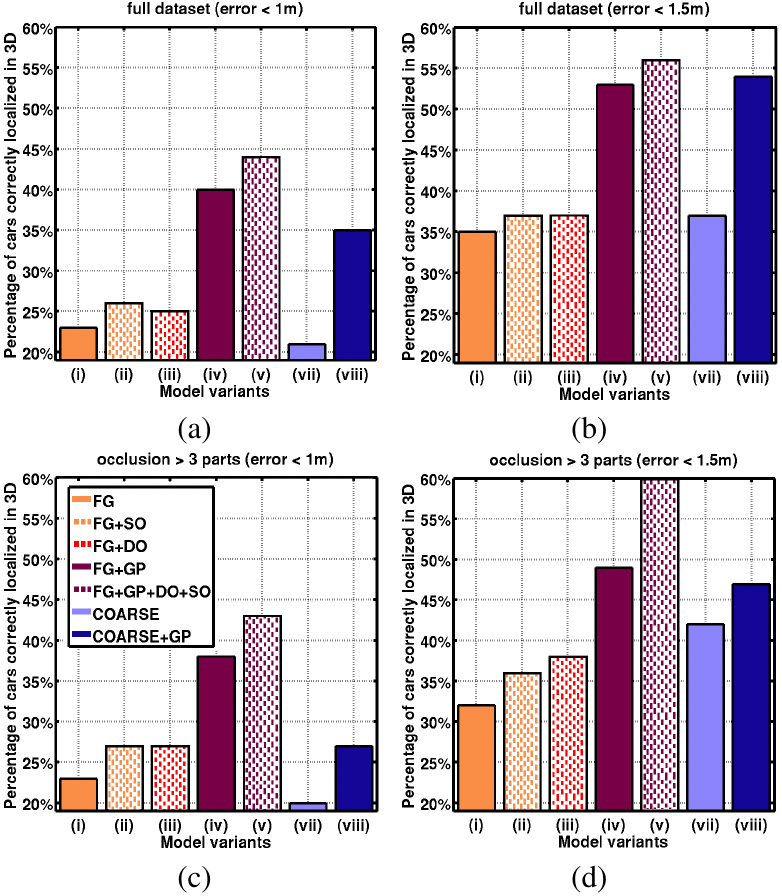}
\end{center}
\caption{3D localization accuracy: percentage of cars correctly
  localized within 1 (a,c) and 1.5 (b,d) meters of ground truth, on
  all (a,b) and occluded (c,d) cars.}
\label{fig:plot_3dloc1}
\end{figure}
%

Having verified that our pre-detection stage is competitive and
provides reasonable object candidates in the image plane, we now
move on to the more challenging task of estimating the 3D location and
pose of objects from monocular images (with known camera intrinsics). As
we will show, the fine-grained representation leads to significant
performance improvements over a standard baseline that considers only
3D bounding boxes, on both tasks.
Our current unoptimized implementation 
takes around $5$ minutes to evaluate the local part detectors in a
sliding-window fashion at multiple scales over the whole image, and
further $20$ minutes per test image for the inference, on a single
core. This is similar to recent deformable face model fitting work,
e.g.~\citet{schoenborn13gcpr}. However, both the sliding-window part
detector and the sample-based inference naturally lend themselves to
massive parallization. In fact the part detector only needs to be
evaluated within the pre-detection bounding boxes, which we do not
exploit at present. Moreover, we set the number of iterations
conservatively, in most cases the results already converge far
earlier.

\subsubsection{3D Object Localization}
\label{sec:3d_localization}
\myparagraph{Protocol.} We measure 3D localization performance by the
fraction of detected object centroids that are correctly localized up
to deviations of $1$, and $1.5$ meters. These thresholds may
seem rather strict for the viewing geometry of KITTI, but in our view
larger tolerances make little sense for cars with dimensions $\approx
4.0\times 1.6$ meters.

In line with existing studies on pose estimation, we base the analysis
on true positive (TP) initializations that meet the PASCAL VOC
criterion for 2D bounding box overlap and whose coarse viewpoint
estimates lie within $45^\circ$ of the ground truth, thus excluding
failures of pre-detection. We perform the analysis for three settings
(Tab.~\ref{tab:results_3d}):~\emph{(i)} over our full testset ($517$
of $982$ TPs);~\emph{(ii)} only over those cars that are partially
occluded, \ie $1$ or more of the parts that are not self-occluded by
the object are not visible ($234$ of $672$ TPs); and \emph{(iii)} only
those cars that are severely occluded, \ie $4$ or more parts are not
visible ($113$ of $476$ TPs). Fig.~\ref{fig:plot_3dloc1} visualizes
selected columns of Tab.~\ref{tab:results_3d} as bar plots to
facilitate the comparison.

\myparagraph{Results.}
In Tab.~\ref{tab:results_3d} and Fig~\ref{fig:plot_3dloc1}, we first
observe that our full system (\fggpdoso, dotted dark red) is the
top performer for all three occlusion settings and both localization
error thresholds, localizing objects with $1$\,m accuracy in $43-44\%$
of the cases and with $1.5$\,m accuracy in $55$--$60\%$ of the
cases. Fig.~\ref{fig:full_vs_coarse} visualizes some examples of our
full system \fggpdoso vs. the stronger baseline \coarsegp.

Second, the basic fine-grained model \fg (orange)
outperforms \coarse (light blue) by $1$--$3\,$ percent points (pp)
corresponding to a relative improvement of $4$--$13\%$ at $1\,$m
accuracy. 
The gains increase by a large margin when adding a ground plane: \fggp
(dark red) outperforms \coarsegp (dark blue) by $5$--$12\,$pp
($13$--$43\%$) at $1\,$m accuracy.
In other words, cars are not 3D boxes. Modeling their detailed shape
and pose yields better scene descriptions, with and without ground
plane constraint.
The results at $1.5\,$m are less clear-cut. It appears that from badly
localized initializations just inside the $1.5\,$m radius, the final
inference sometimes drifts into incorrect local minima outside of
$1.5\,$m.

Third, modeling fine-grained occlusions either independently (\fgso,
dotted orange) or deterministically across multiple objects (\fgdo,
dotted red) brings marked improvements on top of \fg alone. At $1$\,m 
they outperform \fg by $1$--$4\,$pp ($2$--$15\%$) and by $2$--$4\,$pp
($7$--$19\%$), respectively. We get similar improvements at $1.5$\,m,
with \fgso and \fgdo outperforming \fg by $2$--$4\,$pp ($4$--$14\%$),
and $2$--$6\,$pp ($4$--$19\%$) respectively. Not surprisingly,
the performance boost is greater for the occluded cases, and both
occlusion reasoning approaches are in fact beneficial for 3D
reasoning. Fig.~\ref{fig:fggp_vs_full} visualizes some results with
and without occlusion reasoning.

And last, adding the ground plane always boosts the performance for
both the \fg and \coarse models, strongly supporting the case for
joint 3D scene reasoning: at $1\,$m accuracy the gains are
$15$--$18\,$pp ($65$--$81\%$) for \fggp vs. \fg, and 
$7$--$14\,$pp ($30$--$67\%$) for \coarsegp vs. \coarse. Similarly, at
$1.5\,$m accuracy we get $17$--$21\,$pp ($51$--$68\%$) for \fggp
vs. \fg, and $5$--$17\,$pp ($10$--$47\%$) for \coarsegp vs. \coarse.
for qualitative results see Fig.~\ref{fig:fg_vs_fggp}.
%
%

We obtain even richer 3D ``reconstructions'' by replacing wireframes
with nearest neighbors from the database of 3D CAD models
(Fig.~\ref{fig:reconstruct3d}), accurately recognizing hatchbacks (a,
e, f, i, j, l, u), sedans (b, o) and station wagons (d, p, v, w, x),
as well as approximating the van (c, no example in database) by a
station wagon. Specifically, we represent the estimated
wireframe as well as the annotated 3D CAD exemplars as vectors of
corresponding 3D part locations, and find the nearest CAD
exemplar in terms of Euclidean distance, which is then visualized.
Earlier, the same method was used to perform fine-grained object
categorization~\citep{zia13pami}.

\subsubsection{Viewpoint Estimation}
\label{sec:3d_viewpoint}
%
\begin{table*}[t]
\centering
\begin{tabular}{|l|c|c|c|c|c|c|c|c|c|c|c|c|}
\hline
{} & \multicolumn{4}{|c|}{full dataset} & \multicolumn{4}{|c|}{occ \textgreater 0 parts} & \multicolumn{4}{|c|}{occ \textgreater 3 parts}\\
\centering {} & {\textless $5^\circ$} & {\textless $10^\circ$} & {3D err} & {2D err} & {\textless $5^\circ$} & {\textless $10^\circ$} & {3D err} & {2D err} & {\textless $5^\circ$} & {\textless $10^\circ$} & {3D err} & {2D err}\\
\hline
\hline
{\emph{(i)} \fg} & 44\% & {\bf 69\%} & {\bf 5$^\circ$} & {\bf 4$^\circ$} & 41\% & 65\% & $6^\circ$ & $4^\circ$ & 35\% & {\bf 58\%} & {\bf 7$^\circ$} & $5^\circ$ \\ 
{\emph{(ii)} \fgso} & 42\% & 66\% & $6^\circ$ & {\bf 4$^\circ$} & 39\% & 62\% & $6^\circ$ & $4^\circ$ & 33\% & 53\% & $8^\circ$ & $5^\circ$\\ 
{\emph{(iii)} \fgdo} & {\bf 45\%} & 68\% & {\bf 5$^\circ$} & {\bf 4$^\circ$} & 44\% & {\bf 66\%} & $6^\circ$ & $4^\circ$ & 36\% & 56\% & {\bf 7$^\circ$} & {\bf 4$^\circ$}\\ 
{\emph{(iv)} \fggp} & 41\% & 63\% & $6^\circ$ & {\bf 4$^\circ$} & 40\% & 62\% & $6^\circ$ & $4^\circ$ & 36\% & 52\% & $8^\circ$ & $5^\circ$\\ 
{\emph{(v)} \fggpdoso} & {44\%} & {65\%} & $6^\circ$ & {\bf 4$^\circ$} & {\bf 47\%} & {65\%} & {\bf 5$^\circ$} & {\bf 3$^\circ$} & {\bf 44\%} & {55\%} & {\bf $8^\circ$} & {\bf 4$^\circ$}\\ 
\hline
{\emph{(vi)} \cite{zia13cvpr}} &  {-} & {-} & {-} & {$6^\circ$} &  {-} & {-} & {-} & {$6^\circ$} &  {-} & {-} & {-} & {$6^\circ$}\\
\hline
\hline
{\emph{(vii)} \coarse} & 16\% & 38\% & $13^\circ$ & $9^\circ$ & 20\% & 41\% & $13^\circ$ & $6^\circ$ & 21\% & 40\% & $14^\circ$ & $9^\circ$\\ 
{\emph{(viii)} \coarsegp} & 25\% & 51\% & $10^\circ$ & $6^\circ$ & 27\% & 51\% & $10^\circ$ & $5^\circ$ & 23\% & 40\% & $14^\circ$ & $7^\circ$\\ 
\hline
\end{tabular}
\caption{3D viewpoint estimation accuracy (percentage of objects with less than 5$^\circ$ and 10$^\circ$ error) and median angular estimation errors (3D and 2D)}
\label{tab:results_vp}
\end{table*}

%

\begin{figure}[t]
\centering
\includegraphics[width=1.0\columnwidth]{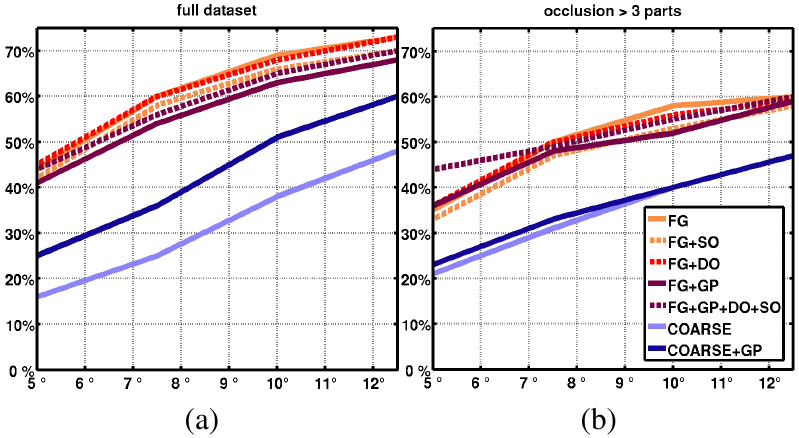}
\caption{Percentage of cars with VP estimation error within $x^\circ$.}
\label{fig:plot_vpacc}
\end{figure}
Beyond 3D location, 3D scene interpretation also requires the
viewpoint of every object, or equivalently its orientation in metric
3D space. Many object classes are elongated, thus their orientation is
valuable at different levels, ranging from low-level tasks such as
detecting occlusions and collisions to high-level ones like enforcing
long-range regularities (\eg cars parked at the roadside are usually
parallel).

\myparagraph{Protocol.}
We can evaluate object orientation (azimuth) in 2D image space as well as in 3D scene space. 2D viewpoint is the apparent azimuth of the object as seen in the image. The actual azimuth relative to a fixed scene direction (called 3D viewpoint), is calculated from the 2D viewpoint estimate and an estimate of 3D object location. We measure viewpoint estimation accuracy in two ways: as the
percentage of detected objects for which the 3D angular error is below
$5^\circ$ or $10^\circ$, and as the median angular error between estimated
and ground truth azimuth angle over detected objects, both in 3D and
2D.

\myparagraph{Results.}
Table~\ref{tab:results_vp} shows the quantitative results, again
comparing our full model and the different variants introduced in
Sec.~\ref{sec:variants}, and distinguishing between the full dataset
and two subsets with different degrees of occlusion. In
Fig.~\ref{fig:plot_vpacc} we plot the percentage of cars whose poses
are estimated correctly up to different error thresholds, using the
same color coding as Fig.~\ref{fig:plot_3dloc1}.

First, we observe that the full
system \fggpdoso (dotted dark red) outperforms the best coarse
model \coarsegp (dark blue) by significant margins of $19$--$21\,$pp
and $14$--$15\,$pp at $5^\circ$ and $10^\circ$ errors respectively,
improving the median angular error by $4^\circ$--$6^\circ$.

Second, all \fg models (shades of orange and red) deliver quite
reliable viewpoint estimates with smaller differences in performance
($\leq 6\,$pp, or $1^\circ$ median error) for 10$^\circ$ error,
outperforming their respective \coarse counterparts (shades of blue)
by significant margins. Observe the clear grouping of curves in
Fig.~\ref{fig:plot_vpacc}.
However, for the high accuracy regime ($\leq$ 5$^\circ$ error),
the full system \fggpdoso (dotted dark red) delivers the best
performance for both occluded subsets, beating the next best
combination \fgdo (dotted red) by $3$ pp and $8$ pp, respectively.

Third, the ground plane helps considerably for the \coarse models
(shades of blue), improving by $9\,$pp for $\leq$5$^\circ$ error, and
$13\,$pp for $\leq$10$^\circ$ over the full data set. Understandably,
that gain gradually dissolves with increasing occlusion.

And fourth, we observe that in terms of median 2D viewpoint estimation
error, our full system \fggpdoso outperforms the pseudo-3D model
of \citep{zia13cvpr} by $2^\circ$--$3^\circ$, highlighting the benefit
of reasoning in true metric 3D space.
%

\subsection{2D Evaluation}
\label{sec:evaluate2d}
While the objective of this work is to enable accurate localization
and pose estimation in 3D (Sec.~\ref{sec:evaluate3d}), we also present
an analysis of 2D performance (part localization and occlusion
prediction in the image plane), to put the work into context.
Unfortunately, a robust measure to quantify how well the wireframe
model fits the image data requires accurate ground truth 2D locations
of even the occluded parts, which are not available. A measure used
previously in~\citet{zia13cvpr} is 2D part localization accuracy only
evaluated for the visible parts, but we now find it to be biased,
because evaluating the model for just the visible parts leads to high
accuracies on that measure, even if the overall fit is grossly
incorrect. We thus introduce a more robust measure below.
%
\begin{table}[t]
\begin{center}
\footnotesize
\renewcommand{\tabcolsep}{1.5pt}
\begin{tabular}{|l|c|c|c|c|c|c|c|c|c|}
\hline
{} & \multicolumn{2}{|c|}{full dataset} & \multicolumn{2}{|c|}{occ \textgreater 0 parts} & \multicolumn{2}{|c|}{occ \textgreater 3 parts}\\
\centering {} & {occl.} & {\#cars} & {occl.} & {\#cars} & {occl.} & {\#cars} \\ 
\centering {} & {pred.} & {\textgreater 70\%} & {pred.} & {\textgreater 70\%} & {pred.} & {\textgreater 70\%}  \\
\centering {} & {acc.} & {parts} & {acc.} & {parts} & {acc.} & {parts}\\

\hline
{\emph{(i)} \fg} & 82\% & 69\% & 70\% & {\bf 68\%} & 57\% & 43\%\\ 
{\emph{(ii)} \fgso} & 87\% & 66\% & 80\% & 63\% & 77\% & 35\%\\ 
{\emph{(iii)} \fgdo} & 84\% & 70\% & 72\% & 67\% & 62\% & {\bf 47\%}\\ 
{\emph{(iv)} \fggp} & 82\% & 68\% & 68\% & 67\% & 57\% & 46\%\\ 
{\emph{(v)} \fggpdoso} & {\bf 88\%} & {\bf 71\%} & {82\%} & {67\%} & {79\%} & {44\%}\\ 
\hline
{\emph{(vi)} \cite{zia13cvpr}} &  87\% & 64\% & {\bf 84\%} & 61\% & {\bf 84\%} & 32\%\\
\hline
\hline
{\emph{(vii)} \coarse} & {---} & {---} & {---} & {---} & {---} & {---}\\
{\emph{(viii)} \coarsegp} & {---} & {---} & {---} & {---} & {---} & {---}\\
\hline
\end{tabular}
\end{center}
\caption{2D accuracy. Part-level occlusion prediction accuracy and percentage of cars which have \textgreater 70\% parts accurately localized.}
\label{tab:results_2d}
\end{table}
%
%

\myparagraph{Protocol.}
We follow the evaluation protocol commonly applied for human body pose
estimation and evaluate the number of correctly localized parts, using
a relative threshold adjusted to the size of the reprojected car ($20$
pixels for a car of size $500\times 170$ pixels, \ie $\approx 4\%$ of
the total length~\citep[c.f.][]{zia13cvpr}). We use this threshold to
determine the percentage of detected cars for which $70\%$ or more of
all (not self-occluded) parts are localized correctly, evaluated only on
cars for which at least $70\%$ of the (not self-occluded) parts are
visible according to ground truth. We find this measure to be more
robust, since it favours sensible fits of the overall wireframe.

Further, we calculate the percentage of (not self-occluded) parts for
which the correct occlusion label is estimated. For the model variants
which do not use the occluder representation (\fg and \fggp), all
candidate parts are predicted as visible.

\myparagraph{Results.}
Tab.~\ref{tab:results_2d} shows the results for both 2D part
localization and part-level occlusion estimation. We observe that our
full system \fggpdoso is the highest performing variant over the full
data set (88\% part-level occlusion prediction accuracy and 71\% cars
with correct part localization). For the occluded subsets, the full
system performs best among all \fg models on occlusion prediction,
whereas the results for part localization are less conclusive. An
interesting observation is that methods that use 3D context
(\fggpdoso, \fggp, \fgdo) consistently beat (\fgso), \ie inferring
occlusion is more brittle from (missing) image evidence alone than
when supported by 3D scene reasoning.

Comparing the pseudo-3D baseline \citep{zia13cvpr} and its proper
metric 3D counterpart \fgso, we observe that, indeed, metric 3D
improves part localization by $2$--$3$ pp (despite inferior
part-level occlusion prediction). In fact, all \fg variants
outperform \citet{zia13cvpr} in part localization by significant
margins, notably \fggpdoso ($6$--$12$ pp).

On average, we note that there is only a weak (although still positive)
correlation between 2D part localization accuracy and 3D localization
performance (Sec.~\ref{sec:evaluate3d}). In other words, whenever
possible \emph{3D reasoning should be evaluated in 3D space}, rather
than in the 2D projection.\footnote{Note, there is no 3D counterpart
to this part-level evaluation, since we see no way to obtain
sufficiently accurate 3D part annotations.}
%
%
%
%

\begin{figure*}[t]
\centering
\renewcommand{\tabcolsep}{1pt}
\includegraphics[width=2.1\columnwidth]{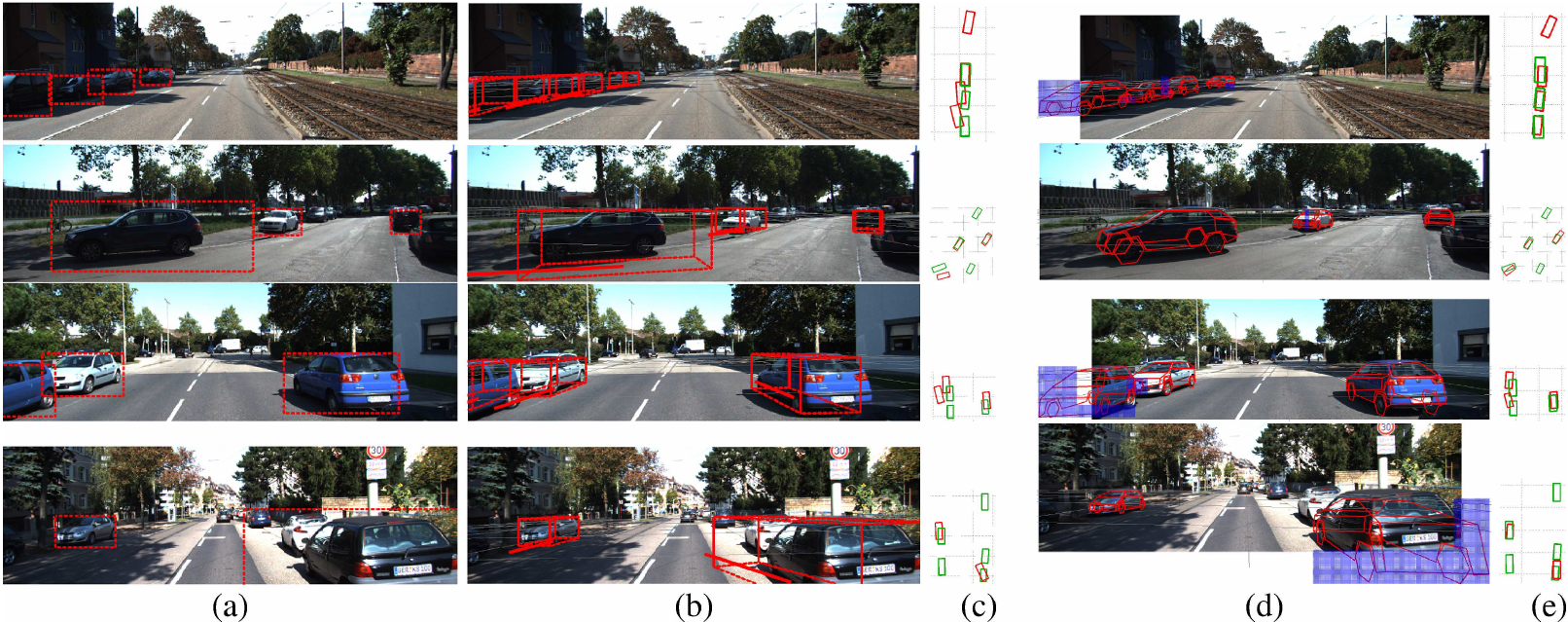}
\caption{\coarsegp (a-c) vs \fggpdoso (d,e). (a) 2D bounding box
detections, (b) \coarsegp based on (a), (c) bird's eye view of
(b), (d) \fggpdoso shape model fits (blue: estimated occlusion masks),
(e) bird's eye view of (d). Estimates in red, ground truth in green.}
\label{fig:full_vs_coarse}
\end{figure*}

\begin{figure*}[t]
\centering
\renewcommand{\tabcolsep}{1pt}
\includegraphics[width=2.1\columnwidth]{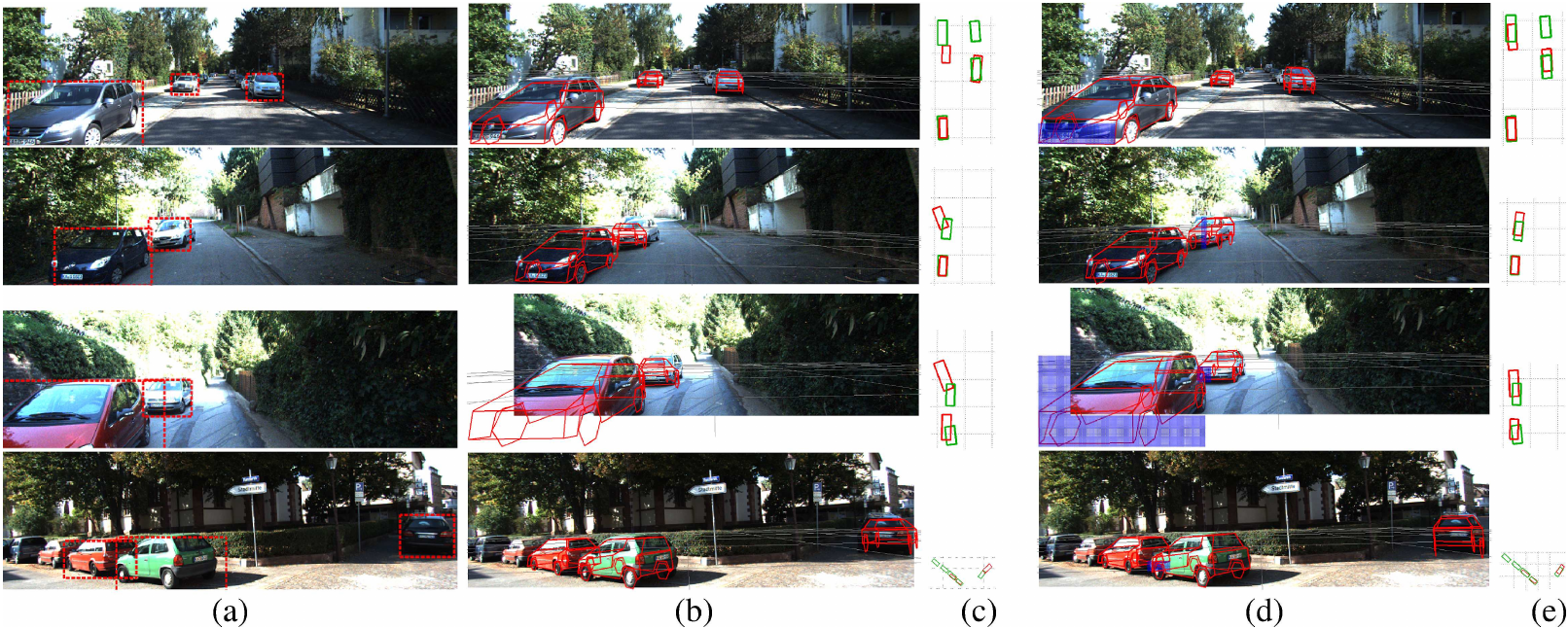}
\caption{\fggp (a-c) vs \fggpdoso (d,e). (a) 2D bounding box
detections, (b) \fggp based on (a), (c) bird's eye view of
(b), (d) \fggpdoso shape model fits (blue: estimated occlusion masks),
(e) bird's eye view of (d). Estimates in red, ground truth in green.}
\label{fig:fggp_vs_full}
\vspace{-0.7em}
\end{figure*}

\begin{figure*}[t]
\centering
\renewcommand{\tabcolsep}{1pt}
\includegraphics[width=2.1\columnwidth]{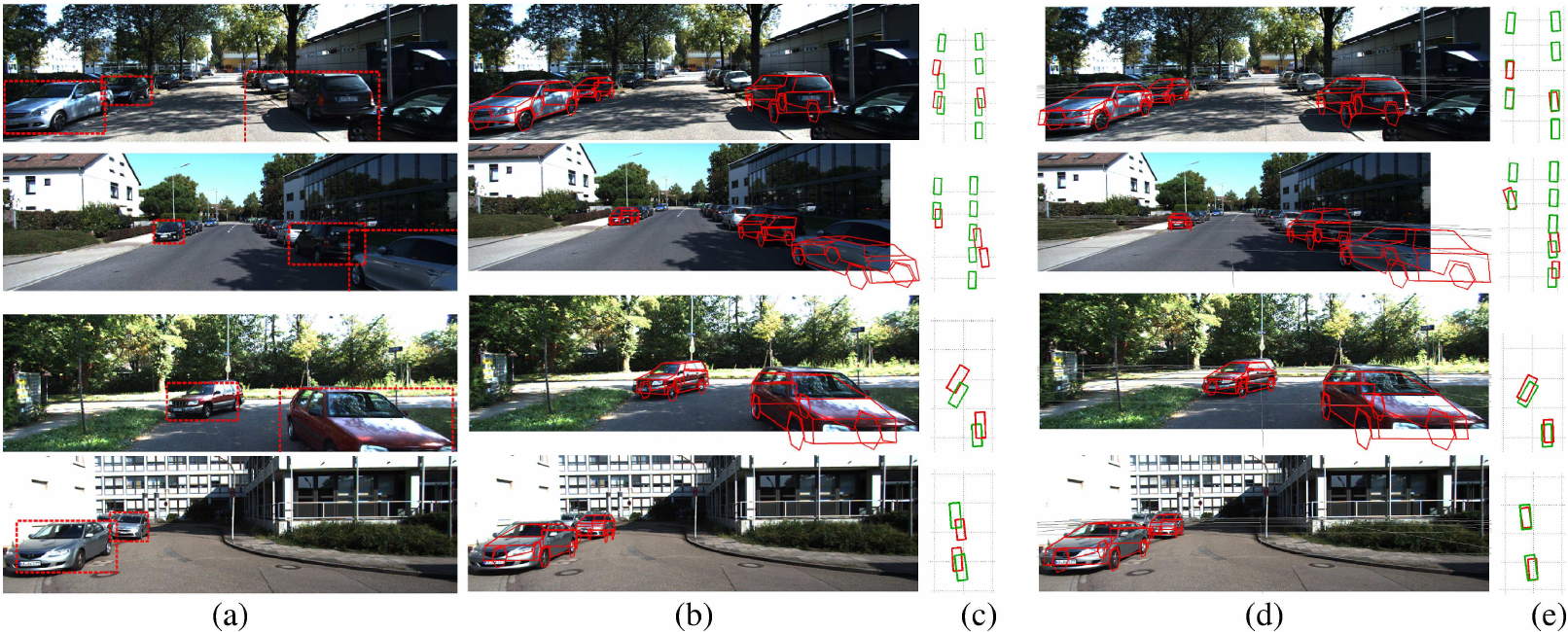}
\caption{\fg (a-c) vs \fggp (d,e). (a) 2D bounding box
detections, (b) \fg based on (a), (c) bird's eye view of
(b), (d) \fggp shape model fits,
(e) bird's eye view of (d). Estimates in red, ground truth in green.
}
\label{fig:fg_vs_fggp}
\vspace{-0.7em}
\end{figure*}

\begin{figure*}[t]
\centering
\renewcommand{\tabcolsep}{1pt}
\includegraphics[width=2.1\columnwidth]{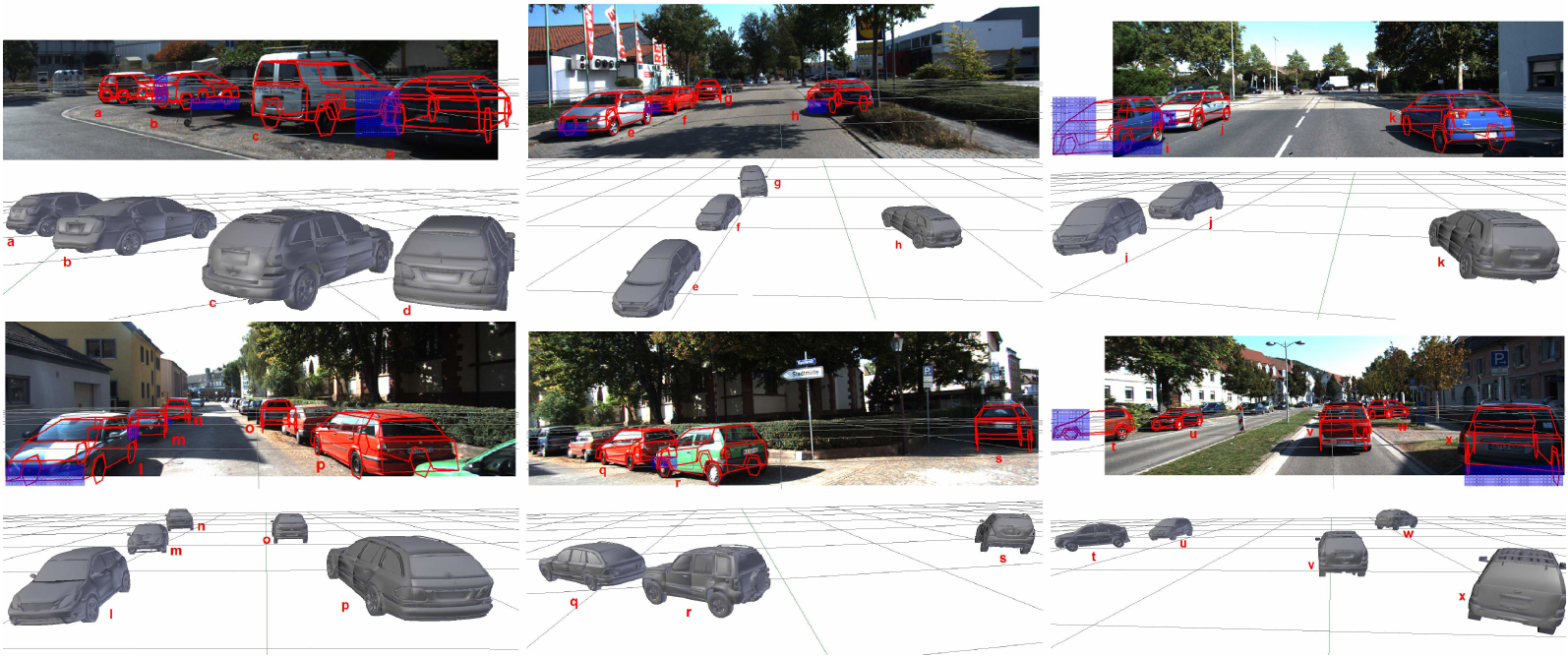}
\caption{Example detections and corresponding 3D reconstructions.}
\label{fig:reconstruct3d}
\end{figure*}
\vspace{-0.51cm}
\section{Conclusion}
\label{sec:conclusion}
We have approached the 3D scene understanding problem from the
perspective of detailed deformable shape and occlusion modeling,
jointly fitting the shapes of multiple objects linked by a common
scene geometry (ground plane).
Our results suggest that detailed representations of object shape are
beneficial for 3D scene reasoning, and fit well with scene-level
constraints between objects.
%
By itself, fitting a detailed, deformable 3D model of cars and
reasoning about occlusions resulted in improvements of $16$--$26\%$ in
object localization accuracy (number of cars localized to within 1m in
3D), over a baseline which just lifts objects' bounding boxes into the
3D scene.
%
Enforcing a common ground plane for all 3D bounding boxes improved
localization by $30$--$67\%$.
When both aspects are combined into a joint model over multiple cars
on a common ground plane, each with its own detailed 3D shape and
pose, we get a striking $104$--$113\%$ improvement in 3D localization
compared to just lifting 2D detections, as well as a reduction of the
median orientation error from $13^\circ$ to $5^\circ$.
We also find that the increased accuracy in 3D scene coordinates is
not reflected in improved 2D localization of the shape model's parts,
supporting our claim that 3D scene understanding should be carried out
(and evaluated) in an explicit 3D representation.

An obvious limitation of the present system, to be addressed in future
work, is that it only includes a single object category, and applies
 to the simple (albeit important) case of scenes with a
dominant ground plane.
In terms of technical approach it woud be desirable to develop a
better and more efficient inference algorithm for the joint scene model.
Finally, the bottleneck where most of the recall is lost is
the 2D pre-detection stage. Hence, either better 2D object detectors
are needed, or 3D scene estimation must be extended to run directly on
entire images without initialization, which will require greatly
increased robustness and efficiency.

\myparagraph{Acknowledgements. }This work has been supported by the Max$\,$Planck$\,$Center$\,$for$\,$Visual$\,$Computing$\,$\&$\,$Communication.
%
\vspace{-0.51cm}
\bibliographystyle{aps-nameyear}      
\bibliography{references}                


\end{document}